\pdfoutput=1
\documentclass[letterpaper,10pt,conference]{ieeeconf}
\IEEEoverridecommandlockouts
\usepackage[utf8]{inputenc}
\usepackage[T1]{fontenc}
\usepackage{times}
\usepackage{amsmath,amssymb,bm}
\usepackage{graphicx,booktabs,cite}
\usepackage{threeparttable}
\usepackage{placeins,float}
\usepackage{dblfloatfix}
\makeatletter
\let\NAT@parse\@undefined
\makeatother
\usepackage[hidelinks]{hyperref}
\newcommand{\R}{\mathbf{R}}
\newcommand{\p}{\mathbf{p}}
\newcommand{\vct}{\mathbf{v}}
\newcommand{\T}{\mathbf{T}}
\newcommand{\Rn}{\mathbb{R}}
\newcommand{\I}{\mathbb{I}}
\newcommand{\bx}{\mathbf{x}}
\newcommand{\bP}{\mathbf{P}}
\newcommand{\bJ}{\mathbf{J}}
\newcommand{\bu}{\mathbf{u}}
\newcommand{\bn}{\mathbf{n}}
\newcommand{\bnu}{\bm{\nu}}
\newcommand{\bd}{\mathbf{d}}
\newcommand{\bQ}{\mathbf{Q}}
\newcommand{\Rz}{\mathbf{R}_z}
\title{\LARGE\bf LP-LIL: Tightly-Coupled LiDAR-Inertial Localization on Prior Maps with Landmark-Pair Global Relocalization}
\author{Xiaoji Niu, Yanlei Zhang, Hailiang Tang, Man Yuan, and Tisheng Zhang%
\thanks{This work was supported in part by the National Natural Science Foundation of China under Grants 42504027 and 42374034, and the Key Research and Development Program of Hubei Province under Grant 2024BAB024. \textit{(Corresponding author: Hailiang Tang.)}}%
\thanks{Tisheng Zhang and Xiaoji Niu are with the GNSS Research Center, Wuhan University, Wuhan 430079, China, and also with the Hubei Luojia Laboratory, Wuhan 430079, China (e-mail: \mbox{zts@whu.edu.cn}; \mbox{xjniu@whu.edu.cn}).}%
\thanks{Yanlei Zhang is with the School of Geodesy and Geomatics and the GNSS Research Center, Wuhan University, Wuhan 430079, China (e-mail: \mbox{yanleizhang@whu.edu.cn}).}%
\thanks{Man Yuan and Hailiang Tang are with the GNSS Research Center, Wuhan University, Wuhan 430079, China (e-mail: \mbox{yuanman@whu.edu.cn}; \mbox{thl@whu.edu.cn}).}%
}
\begin{document}
\flushbottom
\maketitle
\thispagestyle{plain}
\pagestyle{plain}

\begin{abstract}
Prior-map localization is widely used for mobile robot navigation. However, fast and reliable global localization remains challenging at startup without an initial pose estimate. After off-map travel, accumulated drift can prevent local registration from restoring map constraints. This letter proposes LP-LIL, a tightly-coupled LiDAR-inertial localization system for routes with incomplete map coverage. We tightly fuse IMU measurements and scan-to-map measurements using an iterated error-state Kalman filter. When prior-map constraints become insufficient, we switch to local LiDAR-inertial odometry for state estimation and restore map-based localization through landmark-pair relocalization upon re-entry. Our landmark-pair relocalizer combines the relative geometry of landmark pairs with the normals of multiple planes associated with each landmark to reduce matching ambiguity. Extensive experiments in parking lots, building complexes, and urban roads were conducted. On six fully mapped sequences, LP-LIL achieved centimeter-level localization accuracy, reducing the position root mean square error (RMSE) by 40.5\% compared with the loosely-coupled method. Across six sequences containing 13 departures from mapped areas, LP-LIL successfully restored prior-map constraints after every departure. In global relocalization experiments, the landmark-pair relocalizer achieved a higher success rate (from 67.9\% to 95.5\%) with lower storage and processing costs than an adapted triangle-descriptor baseline.
\end{abstract}


\section{Introduction}
\PARstart{R}{eliable} localization is essential for autonomous mobile robots. Visual localization, however, may degrade under poor or changing lighting. LiDAR is largely unaffected by lighting and captures structure accurately, so it is widely used for mobile robot localization~\cite{yin2024survey,chen2020spm}. LiDAR-inertial odometry (LIO) fuses range measurements with inertial measurements and tracks motion without external infrastructure~\cite{shan2020liosam,xu2022fastlio2}. However, the accumulated drift limits long-term navigation because no global reference is available. 

This drift can be limited by matching the current LiDAR scans to a prebuilt point-cloud map~\cite{koide2024prior,liofusion2023}. However, the robot may leave the mapped area, or the map may lack some of the surfaces currently observed by the LiDAR. Under these conditions, unreliable scan-to-map correspondences may lead to erroneous state updates. When prior-map constraints become unreliable, temporary mapping and incremental map management allow local estimation to continue, but the drift still accumulates without a global reference. Over time and distance, the drift may become too large for local registration to restore the map constraints, even after the robot returns to the mapped area.

Global relocalization estimates the pose in the map frame without an accurate initial pose estimate, allowing global initialization at startup and pose recovery after a long departure. Relocalization methods typically use descriptors to retrieve candidate places~\cite{kim2018sc,kim2022scpp,yuan2023std,yuan2024btc}, followed by geometric matching to estimate the pose~\cite{xu2023ringpp,drost2010ppf,yang2021teaser,lim2022quatro,lim2024quatropp,aoki2024bbs}. Once verified, the pose estimate provides the reference in the map frame needed to start or resume map-constrained estimation.

In this letter, we propose LP-LIL for localization across prior-map boundaries. We directly constrain the navigation state using scan-to-map point-to-plane measurements and switch to local LIO when prior-map constraints become insufficient. Verified landmark-pair relocalization and state handoff restore map-based localization upon re-entry. Our main contributions are summarized as follows:
\begin{enumerate}
\item A tightly-coupled LiDAR-inertial localization system is proposed to handle incomplete prior-map coverage. We assess prior-map constraints to guide switching to local LIO and restore map-based localization through verified global relocalization and state handoff.
\item A landmark-pair global relocalization method is proposed to merge corner detections from multiple supporting planes into landmarks while retaining their plane normals. By combining pairwise geometry with plane normals and local geometric features, the method reduces matching ambiguity, while pair construction generates fewer descriptor combinations than triangle-based methods.
\item  Extensive  experiments  were  conducted  on   12  robot sequences in parking  lots, building complexes,  and urban roads, covering localization accuracy, recovery after all 13 departures, and relocalization cost.
\end{enumerate}

\begin{figure*}[!t]
\centering
\includegraphics[width=0.68\textwidth]{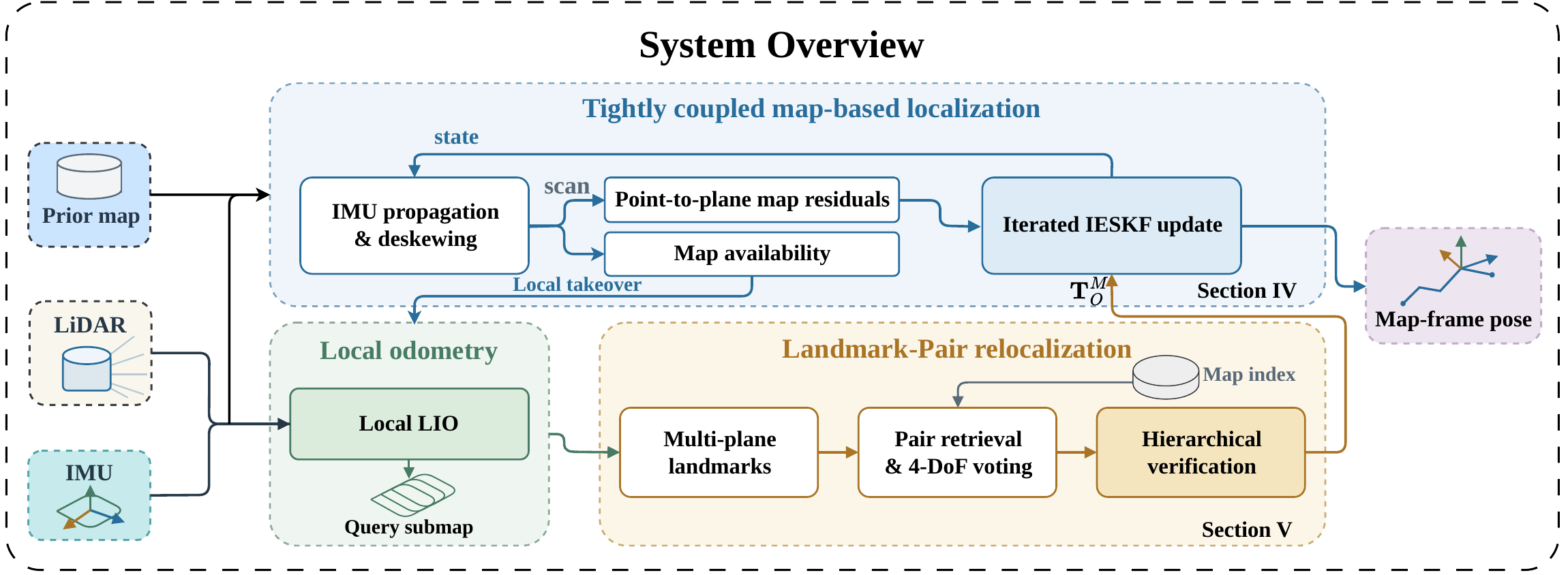}
\caption{Overview of LP-LIL. An IESKF tightly fuses IMU measurements and scan-to-map measurements. Local LIO estimates the state when prior-map constraints become insufficient. The landmark-pair relocalizer provides a verified transformation for initialization and re-entry, after which the state and covariance handoff restores map-constrained estimation.}
\label{fig:system}
\end{figure*}

\section{Related Works}\label{sec:related}

\subsection{LiDAR-Inertial Navigation System}
LiDAR-inertial odometry is a common choice for mobile robot localization. LIO-SAM combines inertial preintegration with LiDAR factors in a factor graph~\cite{shan2020liosam}, whereas FAST-LIO2 uses an iterated filter with direct point matching~\cite{xu2022fastlio2}. FF-LINS introduces frame-to-frame constraints for solid-state LiDAR\cite{tang2023fflins}, and BA-LINS uses bundle adjustment\cite{tang2025balins}. MSC-LIO tracks same-plane clusters\cite{zhang2025msclio}, and SE-LIO relies on plane and cylinder primitives~\cite{zhang2023selio}. Without a global reference, drift still accumulates, motivating the use of prior maps. Koide et al. combine scan-to-scan, scan-to-map, and inertial factors on a prior map, but leave recovery after tracking failure for future work~\cite{koide2024prior}. ROLL maintains a temporary map when global matching becomes unreliable and uses odometry to initialize later registration, but its evaluation assumes a known initial pose~\cite{peng2022roll}. LL-Localizer stores prior-map and incremental geometry in a dynamic i-Octree and uses branch and bound for startup localization~\cite{li2025llloc}. However, temporary mapping and startup localization alone do not ensure recovery after a long departure. LP-LIL addresses this problem through landmark-pair relocalization without an accurate initial pose estimate and verified state handoff, restoring map constraints even when drift prevents local registration.

\subsection{Global Relocalization}
Global relocalization generally involves place recognition and pose estimation~\cite{yin2024survey}. Many descriptors used for place recognition were originally developed for loop closure detection. Scan Context and Scan Context++ encode scans in polar grids~\cite{kim2018sc,kim2022scpp}. The stable triangle descriptor (STD) and the binary and triangle combined (BTC) descriptor use triangle geometry for retrieval and relative pose estimation~\cite{yuan2023std,yuan2024btc}. These methods can estimate relative poses between keyframes, but directly localizing a scan within the entire prior map remains challenging~\cite{yin2024outram,ma2025tripletloc}. RING++ performs place recognition and pose estimation on a sparse scan map and refines the pose with iterative closest point (ICP) registration~\cite{xu2023ringpp}. 3D-BBS searches the pose space directly using branch and bound, with sparse voxel maps and GPU batch processing to reduce memory use and runtime~\cite{aoki2024bbs}. Outram uses semantic scene graphs and global consistency checks for scan-to-map localization, but outlier removal can be costly when many correspondences remain~\cite{yin2024outram}. TripletLoc uses semantic triplet descriptors to limit correspondences and road-surface normals to constrain rotation, but reports lower success rates when few instances are available~\cite{ma2025tripletloc}. These compact semantic maps reduce storage but require semantic segmentation and instance extraction. Outram-GSF~\cite{yin2025outramgsf} models continuous semantic distributions to improve matching in repeated scenes at additional processing cost. LP-LIL combines landmark-pair geometry with normals from multiple supporting planes to reduce matching ambiguity without semantic preprocessing. Pair matching reduces descriptor combinations, while hierarchical verification limits the number of costly registrations. 

\section{System Overview}\label{sec:overview}
Fig.~\ref{fig:system} shows the overall workflow of LP-LIL. The inputs are LiDAR scans, IMU measurements, and a prior point-cloud map. The local LIO first builds a query submap, from which the landmark-pair relocalizer estimates the initial pose. After verification, the iterated error-state Kalman filter (IESKF) fuses IMU measurements and scan-to-map measurements to update the navigation state. When prior-map constraints become insufficient, the last valid state and covariance initialize local LIO, which estimates the state and builds a query submap. When the robot re-enters the mapped area, the relocalizer estimates the transformation from the local odometry frame to the prior-map frame. Once the result is verified, the state and covariance are transformed back into the map frame, and map-constrained estimation resumes.

\section{Map-Availability-Aware Tightly-Coupled Localization}\label{sec:fusion}

\subsection{LiDAR-Inertial State Estimation}\label{sec:update}
LP-LIL fuses IMU measurements and scan-to-map measurements in the IESKF. Each scan-to-map correspondence directly constrains the navigation state. The state consists of the IMU position $\p$, velocity $\vct$, and rotation $\R$ (from the IMU frame to the map frame $\mathcal M$) expressed in $\mathcal M$, the gyroscope and accelerometer biases $\mathbf b_g$ and $\mathbf b_a$ in the IMU frame, and the gravity direction $\bn$ in $\mathcal M$,
\begin{equation}
\bx=(\p,\vct,\R,\mathbf b_g,\mathbf b_a,\bn)
\in\Rn^{12}\times \mathrm{SO}(3)\times \mathbb{S}^2.
\label{eq:state}
\end{equation}
The gravity direction is also estimated to reduce attitude errors arising from inaccurate attitude initialization. The 17-dimensional error state uses right-multiplicative attitude perturbations and a two-dimensional gravity tangent space~\cite{he2021manifold}. For both IMU propagation and scan deskewing, we follow the procedure used in FAST-LIO2~\cite{xu2022fastlio2}.

Point-to-plane residuals are constructed as in FAST-LIO2~\cite{xu2022fastlio2}. Using the state in~\eqref{eq:state}, each deskewed point $\mathbf q_j$ is transformed from the IMU frame to the map frame. For a matched plane with unit normal $\bu_j$ and offset $d_j$, the residual is
\begin{equation}
r_j=\bu_j^\top(\R\mathbf q_j+\p)+d_j.
\label{eq:resid}
\end{equation}
The IESKF combines these residuals with the inertial prediction through
\begin{equation}
\min_{\bx}\;
\|\bx\boxminus\hat{\bx}^{-}\|^2_{(\bP^{-})^{-1}}
+\sum_{j\in\mathcal V}\rho_H(r_j/\sigma_j),
\label{eq:update}
\end{equation}
where $\hat{\bx}^{-}$ and $\bP^{-}$ are the predicted state and covariance, $\sigma_j$ is the residual standard deviation, and $\rho_H$ is the Huber loss. A correspondence is included in $\mathcal V$ only if the planarity, fitting-error, and distance checks are passed. The residuals update the state and covariance, and Huber weighting limits the influence of outliers.

\subsection{Map Constraint Assessment and Local LIO Switching}\label{sec:availability}
LP-LIL evaluates the availability of prior-map constraints using scan-to-map overlap, the valid correspondence count $n_v=|\mathcal V|$, and the smallest eigenvalue of the weighted pose information matrix. Overlap measures the fraction of queried points with nearby map points. Linearization of~\eqref{eq:resid} gives the pose Jacobian $\mathbf h_j=\partial r_j/\partial\delta\boldsymbol\xi$, where $\delta\boldsymbol\xi$ contains position and attitude errors. The map residuals in~\eqref{eq:update} form $\boldsymbol\Lambda_{\mathrm{pose}}=\sum_{j\in\mathcal V}w_j\mathbf h_j^\top\mathbf h_j$. The Huber weights are $w_j=1/\sigma_j^2$ for $|r_j|\le3\sigma_j$ and $w_j=3/(\sigma_j|r_j|)$ otherwise. The matrix excludes the inertial prior. The update is rejected when the minimum eigenvalue falls below the threshold. A scan is marked as unsupported when the overlap is low, the correspondences are too few, or the update fails. Persistent loss of prior-map constraints triggers a switch to local LIO, which then assumes responsibility for state estimation. A separate low-overlap threshold allows faster takeover after a clear departure.

The last reliable map state and covariance initialize local LIO. Buffered LiDAR and IMU measurements are processed to reach the current time. Local LIO estimates the state by matching scans against recent local geometry. We suspend scan-to-map measurement updates and use local scans to build query submaps for relocalization.

\subsection{Map Re-entry and State Handoff}\label{sec:reentry}
During local tracking, LP-LIL checks for map re-entry (i.e., the return to the mapped area) using scan-to-map overlap and prior-map point density. Sampled near-field scan points are transformed using the odometry estimate and the map alignment retained at departure. For each sampled point, the system searches for the nearest prior-map point within a detection radius. The overlap ratio is the fraction of sampled points with a match.

The density check then examines the map around each matched location. The matched map point serves as the center of a neighborhood search that counts prior-map points within a given radius. A match is considered to have sufficient map density when this count reaches the selected threshold. The dense-match fraction is the ratio of dense matches to all matches. Landmark-pair relocalization is triggered when both this fraction and the overlap ratio reach their thresholds and the query submap is ready. The neighborhood radius, point-count threshold, and required dense-match fraction are adjustable parameters.

The detection radius $r_c(d)$ grows linearly with off-map travel distance $d$, up to a chosen maximum. Its growth rate and maximum can be adjusted to suit the expected odometry drift and map geometry. The landmark-pair relocalizer searches around the pose in the prior-map frame predicted by local LIO, with translation and yaw bounds
\begin{equation}
\begin{aligned}
\|\mathbf t-\mathbf t_{\mathrm{prior}}\|&\le \alpha r_c(d),\\
|\psi-\psi_{\mathrm{prior}}|&\le\min(r_{\psi,\max},r_{\psi,0}+k_\psi d).
\end{aligned}
\label{eq:prior}
\end{equation}
Here, $\alpha$ scales the translation search radius. The yaw bound starts at $r_{\psi,0}$, grows at rate $k_\psi$, and is capped at $r_{\psi,\max}$. These bounds are adjustable and do not impose a fixed drift tolerance on the method. Startup localization searches the whole map without this prior. Map constraints resume after relocalization and temporal verification.

A result must pass overlap, registration hit-rate, and mean correspondence-distance checks. Results near the search boundary are not accepted. Repeated estimates must give consistent corrections at the current odometry pose before map-based localization resumes.

After verification, the transformation $\T_{\mathcal O}^{\mathcal M}=(\bQ,\mathbf t)$ maps the local odometry frame $\mathcal O$ to the map frame $\mathcal M$. The navigation state is transformed as
\begin{equation}
\p'=\bQ\p+\mathbf t,\quad
\vct'=\bQ\vct,\quad
\R'=\bQ\R,\quad
\bn'=\bQ\bn.
\label{eq:handoff}
\end{equation}
The biases remain in the IMU frame. The covariance becomes $\bP'=\bJ_T\bP\bJ_T^\top$, where $\bJ_T$ includes the gravity tangent-space transformation. The map-based filter starts from the accepted frame's state and covariance, then processes buffered measurements to reach the current time. If prior-map constraints become insufficient during replay, local LIO remains in control.

\section{Landmark-Pair Global Relocalization}\label{sec:relocalizer}
\begin{figure}[!htbp]
\centering
\includegraphics[width=0.74\columnwidth]{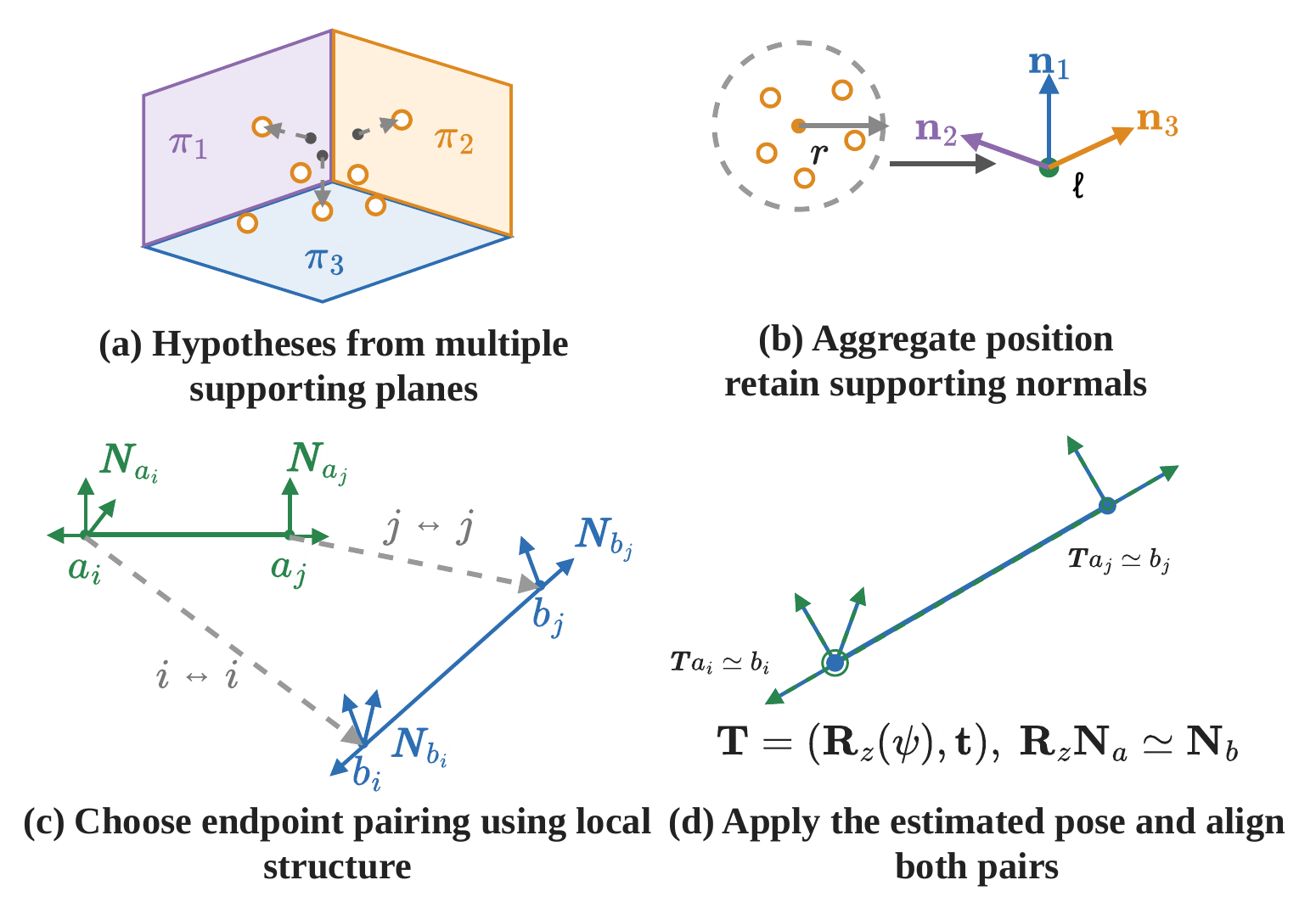}
\caption{Landmark construction and pair matching. (a) The same corner is detected from different supporting planes. (b) Nearby detections are merged within a fixed radius, retaining representative normals from the supporting planes. (c) Supporting-plane information helps select the endpoint correspondence. (d) A matched pair yields a gravity-aligned pose hypothesis. The geometry is illustrated schematically.}
\label{fig:landmarks}
\end{figure}

The landmark-pair relocalizer recovers the pose in the prior-map frame through landmark-pair matching and hierarchical verification. Plane extraction and corner detection follow STD~\cite{yuan2023std}, with nearby detections merged into multi-plane landmarks (Fig.~\ref{fig:landmarks}).

\subsection{Multi-Plane Landmark Aggregation}\label{sec:aggregation}
The same physical corner may be detected several times from the ground, walls, or columns. Keeping these detections as separate keypoints creates duplicates and makes matching sensitive to point density. Nearby detections are therefore merged within a fixed radius, and their supporting-plane information is retained. Each corner hypothesis $h=(\mathbf q_h,r_h,\mathcal N_h)$ contains a position $\mathbf q_h$, a response $r_h$, and supporting-plane normals $\mathcal N_h$. Seeds are selected in descending order of response. Each seed forms $\mathcal C_i$ from the unassigned hypotheses within radius $R_{\mathrm{agg}}$. The cluster is not expanded recursively, so a chain of nearby detections cannot merge different corners. The landmark position is
\begin{equation}
\bm\ell_i=\frac{\sum_{h\in\mathcal C_i}w_h\mathbf q_h}
{\sum_{h\in\mathcal C_i}w_h},\qquad w_h=r_hs_h,
\label{eq:landmark}
\end{equation}
where $s_h$ is the supporting-plane cluster size. This weight reduces the influence of high-response hypotheses whose supporting planes are incomplete. Each landmark keeps the structural attributes
\begin{equation}
\bm\sigma_i=(K_i,\{\bar{\bnu}_{i,k}\},\delta_i^{\mathrm{nd}},\mathrm{PCA}_i),
\label{eq:sigma}
\end{equation}
where $K_i$ counts the retained supporting-plane normals $\mathcal N_i=\{\bar{\bnu}_{i,k}\}$. Similar normals are merged using absolute dot products, regardless of sign. The normal diversity $\delta_i^{\mathrm{nd}}$ is the largest angle $\arccos|\bar{\bnu}_{i,k}^{\top}\bar{\bnu}_{i,l}|$ between retained normals, or zero for a single normal. Neighborhood principal component analysis (PCA) uses the points within 2\,m of the landmark. For covariance eigenvalues $\lambda_1\ge\lambda_2\ge\lambda_3$, the attributes are linearity $(\lambda_1-\lambda_2)/\lambda_1$, planarity $(\lambda_2-\lambda_3)/\lambda_1$, and scattering $\lambda_3/\lambda_1$. Map and query landmark budgets separately control index storage and pair generation.

\subsection{Pair Retrieval and Pose Hypotheses}\label{sec:pair}
A triangle descriptor combines an anchor with two neighbors, producing $k(k-1)/2$ combinations before geometric filtering and duplicate removal~\cite{yuan2023std}. A triangle cannot be matched if one of its three corners is missing. A pair descriptor requires only two landmarks. For an anchor with $k$ neighbors in the search range, the pair descriptor forms $k$ pairs by connecting the anchor to each neighbor. Supporting-plane information helps distinguish pairs with similar lengths in repeated structures. For $(\bm\ell_i,\bm\ell_j)$, the descriptor contains the edge length $l_{ij}=\|\bm\ell_j-\bm\ell_i\|$, the height difference $h_{ij}=|\zeta_j-\zeta_i|$ with $\zeta_i$ the landmark height, and the normal-alignment terms
\begin{equation}
\begin{aligned}
a_i&=\max_{\bnu\in\mathcal N_i}|\bnu^\top\bd_{ij}|,\quad
a_j=\max_{\bnu\in\mathcal N_j}|\bnu^\top\bd_{ij}|,\\
c_{ij}&=|\bnu_i^\top\bnu_j|,
\end{aligned}
\label{eq:pair}
\end{equation}
where $\bd_{ij}$ is the unit edge direction. Edge alignment uses all retained supporting-plane normals, whereas $c_{ij}$ uses the supporting-plane normal of the highest-response seed hypothesis at each endpoint, denoted by $\bnu_i$ and $\bnu_j$, as a lightweight consistency term prior to full multi-normal endpoint scoring. The descriptor also stores $\bm\sigma_i$ and $\bm\sigma_j$. Map and query submaps are gravity-aligned. The map frame is right-handed, with an upward $z$ axis. Under this alignment, the descriptor varies little with yaw or translation.

Map pairs are indexed in advance by the quantized key
\begin{equation}
\kappa_{ij}=(\lfloor l_{ij}/\Delta l\rfloor,\lfloor h_{ij}/\Delta h\rfloor),
\label{eq:index}
\end{equation}
where $\Delta l$ and $\Delta h$ are the quantization steps. Each query pair searches the hash buckets within the descriptor tolerance. Map pairs are assigned to 20\,m $\times$ 20\,m $\times$ 3\,m cells by their midpoints. Matches from different cells are interleaved within each bucket, with regional candidates retained alongside the best matches to prevent one repeated region from consuming the search budget.

Each pair has two possible endpoint correspondences. After the lookup in~\eqref{eq:index}, the normal-edge terms in~\eqref{eq:pair} are compared. For each correspondence, the query normals are rotated by the implied yaw. Endpoint similarity combines support-count compatibility, normal agreement, normal-diversity similarity, and PCA similarity with weights 0.30, 0.40, 0.15, and 0.15. Support counts are grouped as zero, one, or multiple planes, with compatibility scores 1, 0.55, and 0.25 for equal, adjacent, and opposite classes. Normal agreement is the mean, over query normals, of the largest absolute dot product with a map normal. Diversity similarity is $\exp[-(\delta_s^{\mathrm{nd}}-\delta_m^{\mathrm{nd}})^2/(2\times0.25^2)]$, with angles in radians. PCA similarity is $\exp(-D/0.75)$, where $D$ sums the absolute differences in linearity, planarity, and scattering. The correspondence with the higher mean endpoint similarity is retained. Here and in~\eqref{eq:pair}--\eqref{eq:pose}, superscripts $s$ and $m$ denote the query submap and the prior map, respectively. For query endpoints $i,j$ matched to map endpoints $a,b$, let $\bm\Delta^s=\bm\ell_j^s-\bm\ell_i^s$ and $\bm\Delta^m=\bm\ell_b^m-\bm\ell_a^m$. The closed-form 4-DoF pose hypothesis is
\begin{equation}
\begin{aligned}
\psi&=\operatorname{atan2}(\Delta_y^m,\Delta_x^m)
-\operatorname{atan2}(\Delta_y^s,\Delta_x^s),\\
\mathbf t_{xy}&=\bm\ell_{a,xy}^m-(\Rz(\psi)\bm\ell_i^s)_{xy},\\
t_z&=\tfrac12[\zeta_a^m-z(\Rz(\psi)\bm\ell_i^s)
+\zeta_b^m-z(\Rz(\psi)\bm\ell_j^s)].
\end{aligned}
\label{eq:pose}
\end{equation}
Here, $z(\cdot)$ extracts the vertical component. Because matched pairs satisfy tight edge-length tolerances and are averaged during subsequent pose clustering, anchoring $\mathbf t_{xy}$ at one endpoint yields nearly the same horizontal translation as two-endpoint averaging, whereas averaging $t_z$ directly mitigates per-endpoint height noise that does not enter the yaw estimate. Nearly vertical pairs are excluded because their short horizontal baselines provide insufficient yaw information. The hypotheses from~\eqref{eq:pose} are clustered in a discretized $(x,y,z,\psi)$ space. Each cluster stores the mean translation, circular mean yaw, and number of different supporting query landmarks. Counting distinct supporting landmarks limits repeated votes from the same query endpoint.

\subsection{Hierarchical Verification and Acceptance}\label{sec:verification}
Repeated structures can produce many false pose hypotheses. Registering every candidate would be too costly, so candidates are checked in stages. A coarse occupancy map first ranks poses by the fraction of transformed query points in occupied map voxels. A sparse distance field then scores sampled points using hit rates within 0.5 and 1.0\,m and penalties for the mean and 90th-percentile distances. The retained poses undergo a bounded 4-DoF perturbation search, which favors high hit rates and short distances. Poses in the same $(x,y,z,\psi)$ cell form a basin, using 1\,m horizontal, 0.8\,m vertical, and $5^\circ$ yaw bins. Basin representatives are ranked by a weighted score of nearest-map hit rates within 0.5, 1.0, and 1.5\,m and mean and 90th-percentile distances, and the best pose in each basin is retained, avoiding repeated refinement around the same solution. Selected representatives are refined locally and registered using multiresolution generalized ICP (GICP)~\cite{segal2009gicp} to obtain 6-DoF poses. To bound verification cost, coarse screening retains up to $K_c$ candidates---allocating two-thirds to the top distance-field scores, one-sixth to the top voting scores, and one-sixth to spatially diverse cells---followed by up to $K_f$ candidates after 4-DoF screening and up to $K_r$ basin representatives for 6-DoF refinement. This limits registration to distinct hypotheses with sufficient support. For a refined pose $\T^\star$, geometric agreement is measured by correspondence distances and the sampled-point hit rate
\begin{equation}
s_\epsilon(\T^\star)=\frac1{|\mathcal Q|}\sum_{\mathbf q\in\mathcal Q}
\I[d_{\mathcal M}(\T^\star\mathbf q)<\epsilon],
\label{eq:score}
\end{equation}
where $\mathcal Q$ is the sampled query point set, $\I[\cdot]$ equals one when the condition holds and zero otherwise, $d_{\mathcal M}$ is the distance to the nearest map point, and $\epsilon$ is the hit threshold. The hit rate in~\eqref{eq:score}, mean correspondence distance, and temporal agreement are checked using the conditions in Section~\ref{sec:reentry}. A frame without candidates leaves the confirmation count unchanged. Contradictory evidence resets the count. Once a pose is confirmed, the state handoff in Section~\ref{sec:reentry} restores map-constrained estimation.

\section{Experiments and Results}\label{sec:experiments}
12 sequences were collected with the four-wheeled robot described in~\cite{tang2025i2navrobot}, covering multilevel parking lots, building complexes, and urban roads (Table~\ref{tab:data}).

\subsection{Experimental Setup}\label{sec:exp-setup}

\subsubsection{Platform and Data}
Localization used only a Livox Mid-360 LiDAR and an ADIS16465 IMU mounted on the robot. A Leador POS-A15 inertial navigation system (INS) and a NovAtel OEM719 receiver provided the reference trajectories, following the procedure in~\cite{tang2025i2navrobot}. During off-map travel, the high-grade INS maintained the reference trajectory with sufficient accuracy to evaluate localization recovery at re-entry. Trajectory errors were evaluated with evo~\cite{grupp2017evo}. The data were replayed on a computer with an Intel Core Ultra 7 265K processor, 32\,GB of memory, and Ubuntu 20.04.

\begin{table}[!htbp]
\centering
\renewcommand{\arraystretch}{0.90}
\setlength{\aboverulesep}{1pt}
\setlength{\belowrulesep}{1.5pt}
\begin{threeparttable}
\caption{Robot sequences and prior maps}
\label{tab:data}
\footnotesize
\setlength{\tabcolsep}{4pt}
\begin{tabular}{cccccc}
\toprule
Seq. & Type & Scene & Map & Time (s) & Dist. (m) \\
\midrule
P00 & In-map & Multilevel parking & M1 & 1141 & 1583 \\
P01 & In-map & Multilevel parking & M1 & 1060 & 1224 \\
P02 & In-map & Multilevel parking & M1 & 1257 & 1442 \\
B00 & In-map & Buildings & M3 & 1209 & 1675 \\
B01 & In-map & Buildings & M2 & 1369 & 1730 \\
B03 & In-map & Trees/buildings & M4 & 1237 & 1678 \\
B02 & Holes & Trees/buildings & M4 & 1757 & 2522 \\
B04 & Holes & Trees/buildings & M4 & 1300 & 1398 \\
L01 & Staged & Site roads & M3 & 1728 & 1739 \\
L02 & Traverse & Urban roads & M5 & 2000 & 2138 \\
L03 & Traverse & Urban roads & M5 & 2084 & 2309 \\
L04 & Traverse & Urban roads & M5 & 1789 & 2086 \\
\bottomrule
\end{tabular}
\begin{tablenotes}[flushleft]
\footnotesize
\item[] “Holes” denotes sequences with missing mapped surfaces, “Staged” denotes intermittent departures from the mapped area, and “Traverse” denotes long off-map traverses.
\end{tablenotes}
\end{threeparttable}
\end{table}

\begin{figure*}[!tbp]
\centering
\includegraphics[width=0.68\textwidth]{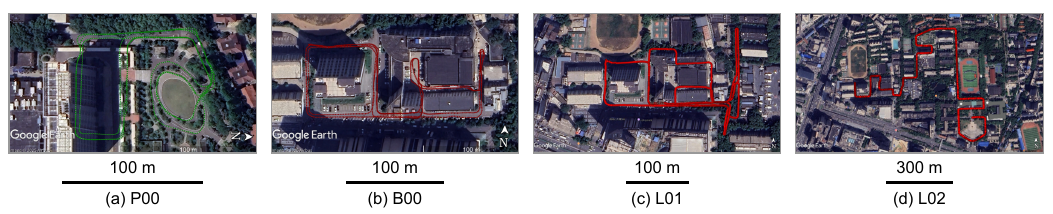}
\caption{Representative collection routes. (a) P00 in a multilevel parking lot. (b) B00 around buildings. (c) L01 along site roads with three staged departures. (d) L02 with two long off-map traverses. Scale bars indicate the corresponding ground distances.}
\label{fig:scenes}
\end{figure*}

Five prior maps (M1--M5) were built using professional commercial SLAM devices. M1 contains a plaza and two underground parking levels with overlapping horizontal projections. M2 and M3 are two map versions of the same site. The sequence prefixes P, B, and L denote parking, building-area, and road-route datasets, respectively. B02 and B04 include areas with missing mapped surfaces. L01 contains three staged departures. L02--L04 each contain two long off-map excursions (traverses) (Fig.~\ref{fig:scenes}).

\subsubsection{Comparison Methods}
The four baselines were built on the same FAST-LIO2 core~\cite{xu2022fastlio2} because reliable open-source implementations of the prior-map methods considered here were unavailable. All methods used the same extrinsics, initialization, and maps. FAST-LIO2 uses a local map without prior-map constraints. Map-IEKF uses the map-update core of LP-LIL but omits the checks of map constraints. Loosely-coupled localization (Loose) uses a separately registered map pose as a filter measurement and continues the odometry prediction if registration fails. L-NDT uses the LIO prediction to initialize local registration with the normal distributions transform (NDT) and ICP. Map localization resumes after the result passes an acceptance check~\cite{magnusson2009ndt}.

\subsubsection{Evaluation Metrics and Parameters}
Position accuracy was measured as the root mean square error (RMSE) of the three-dimensional position errors, and attitude accuracy as the RMSE of the relative rotation angles. Evaluation used a 20\,ms timestamp tolerance and fixed-scale rigid alignment for the comparison methods.

Map coverage was classified using the reference trajectory and prior-map geometry. Horizontal map distances defined three intervals: in-map (at most 1\,m), boundary (above 1\,m up to 5\,m), and off-map (above 5\,m). Map neighbors were restricted to within 2\,m in the vertical direction. Off-map intervals separated by at most 30\,s were merged, and intervals shorter than 20\,s were excluded. Recovery was considered successful when the position RMSE returned to the centimeter level after re-entry. Position RMSE was measured over 5--25\,s after the recorded restoration of map constraints, using the same windows for all methods. For the experiments, the re-entry overlap threshold was 80\%. The refined pose required a hit rate above 60\% within 0.5\,m and a mean correspondence distance below 0.25\,m. Three successive corrections had to agree within 0.5\,m and 0.1\,rad. L04 used a hit-rate threshold of 50\% to enable earlier switching at a sparse re-entry boundary dominated by low vegetation and foliage; 60\% would still permit switching farther inside the map. The map landmark limit was 5000 and was increased to 10,000 for the composite M5.

\subsection{Evaluation of the In-Map Localization}\label{sec:exp-inmap}
Table~\ref{tab:main}(a) compares the methods on the six fully mapped sequences. LP-LIL achieved position RMSEs of 0.050--0.117\,m. The RMSE over 72,001 positions was 0.077\,m, compared with 0.129\,m for Loose, a reduction of 40.5\%. Map-IEKF achieved a similar RMSE of 0.076\,m, as both methods share the same map-update core. Their direct use of scan-to-map point-to-plane measurements helps explain this advantage over Loose. FAST-LIO2 achieved RMSEs of 5.665\,m and 12.478\,m on B00 and B03, compared with 0.059\,m and 0.072\,m for LP-LIL. L-NDT reached 18.195\,m on B01, showing that local registration can lose alignment even inside the map when the initial prediction has drifted. LP-LIL had the lowest RMSE on five of the six sequences, while Map-IEKF performed best on B03.
\begin{table}[!htbp]
\centering
\renewcommand{\arraystretch}{0.90}
\setlength{\aboverulesep}{1pt}
\setlength{\belowrulesep}{1.5pt}
\begin{threeparttable}
\caption{Position RMSE (m)}
\label{tab:main}
\footnotesize
\setlength{\tabcolsep}{2pt}
\begin{tabular*}{\columnwidth}{@{\extracolsep{\fill}}cccccc@{}}
\toprule
\multicolumn{6}{@{}c}{\textit{(a) In-map position RMSE (m)}} \\
\midrule
Seq. & FAST-LIO2 & Map-IEKF & Loose & L-NDT & LP-LIL \\
\midrule
P00 & 0.976 & 0.119 & 0.132 & 0.137 & \textbf{0.117} \\
P01 & 0.833 & 0.069 & 0.083 & 0.310 & \textbf{0.068} \\
P02 & 0.293 & 0.084 & 0.093 & 0.324 & \textbf{0.082} \\
B00 & 5.665 & 0.063 & 0.148 & 0.134 & \textbf{0.059} \\
B01 & 1.587 & 0.054 & 0.097 & 18.195 & \textbf{0.050} \\
B03 & 12.478 & 0.045 & 0.187 & 0.154 & \textbf{0.072} \\
\bottomrule
\end{tabular*}
\par\vspace{3pt}\noindent
\begin{tabular*}{\columnwidth}{@{\extracolsep{\fill}}cccccccc@{}}
\toprule
\multicolumn{8}{@{}c}{\textit{(b) After restoration of map constraints}} \\
\midrule
Seq. & $N$ & $t$ (s) & $d$ (m) & Map-IEKF & Loose & L-NDT & LP-LIL \\
\midrule
B02 & 2 & 51 & 74 & 0.090 & 0.234 & 0.171 & \textbf{0.072} \\
B04 & 2 & 104 & 129 & $1.119\times10^6$ & 0.127 & 0.099 & \textbf{0.028} \\
L01 & 3 & 366 & 417 & $2.742\times10^6$ & 0.166 & 0.147 & \textbf{0.027} \\
L02 & 2 & 493 & 578 & $1.559\times10^6$ & 22.068 & 9.083 & \textbf{0.023} \\
L03 & 2 & 348 & 474 & $3.644\times10^6$ & 33.578 & 28.533 & \textbf{0.026} \\
L04 & 2 & 325 & 455 & $2.984\times10^6$ & 47.641 & 48.834 & \textbf{0.052} \\
\bottomrule
\end{tabular*}
\begin{tablenotes}[flushleft]
\footnotesize
\item[] (a) Results on the six fully mapped sequences. (b) $N$ is the number of departures, and $t$ and $d$ are the total off-map time and distance. RMSE values use shared 5--25\,s post-restoration windows, pooled within each sequence. LP-LIL is in bold.
\end{tablenotes}
\end{threeparttable}
\end{table}

\subsection{Evaluation of the Departure and Restoration}\label{sec:exp-offmap}
The six remaining sequences contained 13 departures, and LP-LIL restored map constraints after every one of them. Table~\ref{tab:main}(b) compares position RMSE over the same post-restoration windows. LP-LIL achieved RMSEs of 0.015--0.096\,m across the individual episodes and an RMSE of 0.041\,m over 2600 positions.

\begin{figure}[!htbp]
\centering
\includegraphics[width=0.68\columnwidth]{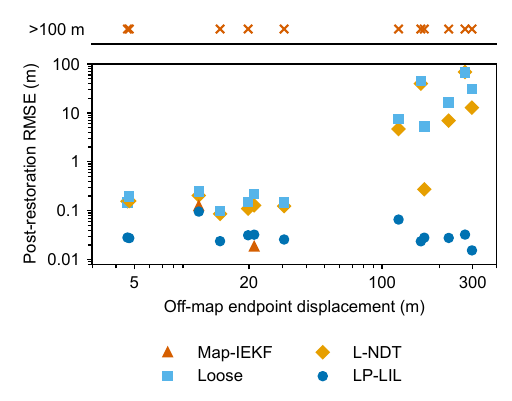}
\caption{Position RMSE after all 13 recorded LP-LIL restorations. All methods are evaluated over the same 5--25\,s window after the recorded LP-LIL restoration of map constraints. The horizontal axis shows the reference endpoint displacement during the departure. Both axes are logarithmic. Errors above 100\,m are shown in a separate upper strip.}
\label{fig:capture}
\end{figure}

Fig.~\ref{fig:capture} shows centimeter-level post-restoration RMSEs across all 13 recovery cases.

\begin{figure*}[!tbp]
\centering
\includegraphics[width=0.64\textwidth,trim=0 3bp 0 5bp,clip]{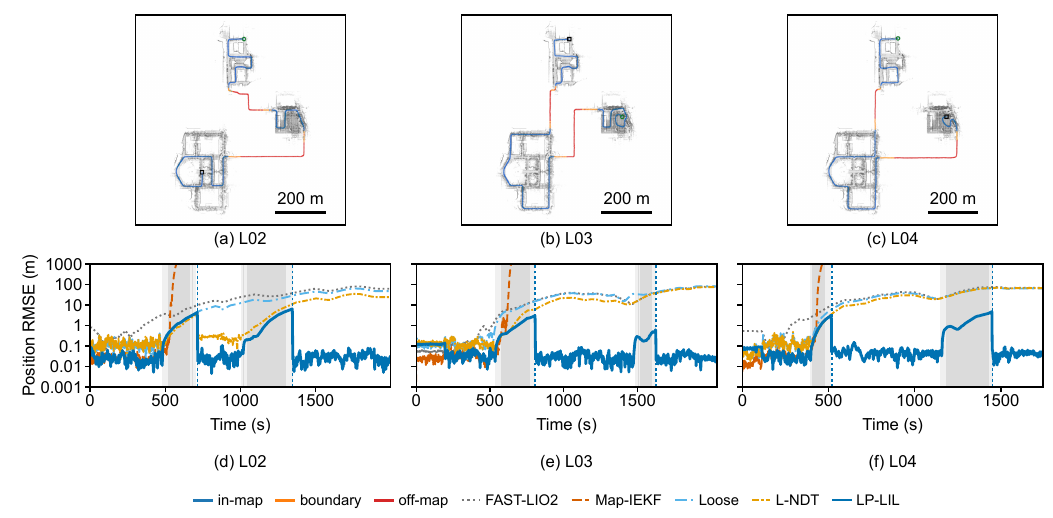}
\caption{Map coverage and localization through long departures. (a)--(c) Routes of L02--L04 overlaid on the prior maps. Blue, orange, and red indicate in-map, boundary, and off-map intervals. (d)--(f) Position RMSE in nonoverlapping 1\,s bins, with the off-map intervals shaded. Vertical blue lines mark the confirmed restoration of map constraints by LP-LIL. The vertical axis is logarithmic and spans 0.001--1000\,m.}
\label{fig:traverse}
\end{figure*}

\subsubsection{Short Departures and Map Holes}
On B04 and L01, LP-LIL achieved post-restoration RMSEs of 0.024--0.031\,m and 0.026--0.028\,m, respectively. Map-IEKF diverged after the map constraints became unavailable. For the two B02 map holes, LP-LIL achieved 0.032\,m and 0.096\,m, whereas Map-IEKF maintained an RMSE of 0.090\,m over the same windows. These results indicate that evaluating prior-map constraints helps avoid erroneous state updates, while small map holes do not always require switching to local LIO.

\subsubsection{Long Traverses}
The six traverses in L02--L04 lasted 75--300\,s and had net displacements of 122--297\,m. LP-LIL restored map constraints after every traverse. Post-restoration RMSEs were 0.015--0.066\,m for LP-LIL and 5.244--66.946\,m for Loose. L-NDT achieved 0.274\,m after the first L02 traverse but 12.842\,m after the second. Fig.~\ref{fig:traverse} shows how the error decreases after re-entry. The landmark-pair search provides a fresh initial pose when the accumulated drift prevents local registration from converging.

\subsection{Evaluation of the Global Relocalization}\label{sec:exp-frontend}
The relocalizer was evaluated separately on 12,294 query frames from seven sequences, and Table~\ref{tab:frontend} reports the success rate for the 11,803 in-map frames among them.

\subsubsection{Evaluation Settings}
Each query submap contained the target keyframe and five neighboring keyframes on each side. Reference poses were taken from the LP-LIL in-map trajectory rather than the external INS/GNSS reference, which is not natively expressed in the prior-map frame and would require additional frame alignment. Because the map-constrained LP-LIL trajectory directly provides native map-frame poses with sub-decimeter accuracy (0.077\,m overall RMSE in Section~\ref{sec:exp-inmap}), its residual error is negligible relative to the 0.5\,m and $1^\circ$ success thresholds used identically for both relocalizers; a query was deemed successful if its final position and rotation errors fell below these thresholds. STD-adapt used the plane-supported corners and triangles of STD~\cite{yuan2023std}. Non-maximum suppression, regional quotas, and neighborhood sizes were adjusted for map-to-submap matching. Three-vertex occupancy and corner-coincidence checks were also applied. STD-adapt used map budgets of 12,000 corners in parking lots and 8000 around buildings. LP-LIL used 5000 map landmarks and 120 query landmarks. Both relocalizers used the same screening and refinement steps. B00 used the newer, more accurate M3 for continuous localization. Global initialization used M2 to cover the underground garage omitted from M3, despite a local 0.2\,m error in M2.

\begin{table}[!htbp]
\centering
\renewcommand{\arraystretch}{0.90}
\setlength{\aboverulesep}{1pt}
\setlength{\belowrulesep}{1.5pt}
\begin{threeparttable}
\caption{Global relocalization success (\%)}
\label{tab:frontend}
\footnotesize
\setlength{\tabcolsep}{7pt}
\begin{tabular}{cccc}
\toprule
Seq. & In-map queries & STD-adapt & LP-LIL \\
\midrule
B00 & 1613 & 91.3 & 99.6 \\
B01 & 1666 & 77.7 & 99.1 \\
P00 & 1379 & 46.0 & 94.9 \\
P01 & 1134 & 20.1 & 96.0 \\
P02 & 1424 & 30.5 & 90.9 \\
B03 & 2195 & 85.1 & 96.5 \\
B02 & 2392 & 87.2 & 92.3 \\
\bottomrule
\end{tabular}
\begin{tablenotes}[flushleft]
\footnotesize
\item[] Results are computed on the 11,803 in-map frames out of the 12,294 evaluated ones. Both methods share the same query submaps, screening, and refinement steps.
\end{tablenotes}
\end{threeparttable}
\end{table}

\subsubsection{Pose Recovery}
LP-LIL achieved an overall success rate of 95.5\%, compared with 67.9\% for STD-adapt (Table~\ref{tab:frontend}). The largest gains occurred in multilevel parking lots, where repeated columns and similar layouts produce ambiguous triangles that compete for the candidate budget. STD requires three corresponding corners, whereas the pair descriptor uses two landmarks and supporting-plane information to reject ambiguous matches.

\subsubsection{Candidate Retrieval and Screening}
\begin{figure}[!htbp]
\centering
\includegraphics[width=0.70\columnwidth]{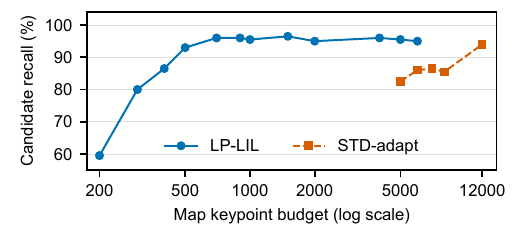}
\caption{Candidate-level recall on 200 fixed M1 queries. A query counts as recalled if any candidate pose is within 2.5\,m and $15^\circ$ of the reference. LP-LIL candidates are taken after pose voting, before FastP2P screening. STD-adapt candidates are taken after coarse verification, before pose clustering.}
\label{fig:frontend}
\end{figure}

Fig.~\ref{fig:frontend} compares candidate-level recall on 200 fixed M1 queries. LP-LIL recall rose from 59.5\% at 200 landmarks to 96.0\% at 700 and ranged from 95.0\% to 96.5\% at higher budgets. With 5000 map keypoints, LP-LIL and STD-adapt achieved 95.5\% and 82.5\% recall, respectively. STD-adapt reached 94.0\% with 12,000 corners.

With candidate budgets set to $K_c=768$ ($512+128+128$), $K_f=300$, and $K_r=3$, screening removed 99.6\% of the raw candidates over all 12,294 queries before basin verification (Table~\ref{tab:screening}). The fraction of queries retaining a candidate within 5\,m and $10^\circ$ of the reference decreased from 98.2\% to 96.7\%, reducing the verification load while preserving useful hypotheses.

\begin{table}[!htbp]
\centering
\renewcommand{\arraystretch}{0.90}
\setlength{\aboverulesep}{1pt}
\setlength{\belowrulesep}{1.5pt}
\begin{threeparttable}
\caption{LP-LIL candidate screening}
\label{tab:screening}
\footnotesize
\setlength{\tabcolsep}{3pt}
\begin{tabular*}{\columnwidth}{@{\extracolsep{\fill}}ccc@{}}
\toprule
Stage & \shortstack{Mean candidates\\per query} & \shortstack{Queries with a\\near-reference pose} \\
\midrule
Raw voting & 81,607 & 98.2\% \\
Coarse screening & 768 & 97.3\% \\
4-DoF screening & 300 & 96.7\% \\
\bottomrule
\end{tabular*}
\begin{tablenotes}[flushleft]
\footnotesize
\item[] Results are computed over all 12,294 queries. A near-reference pose has a position error of at most 5\,m and a rotation error of at most $10^\circ$. The last column gives the percentage of queries containing at least one such candidate. Mean counts are rounded to the nearest integer.
\end{tablenotes}
\end{threeparttable}
\end{table}

\subsubsection{Relocalization Efficiency}\label{sec:exp-efficiency}
Table~\ref{tab:efficiency} compares the storage and query processing costs of the two relocalizers. Pair indexing and regional quotas reduce the candidates sent to the shared refinement backend, lowering storage use and processing time.
\begin{table}[!htbp]
\centering
\renewcommand{\arraystretch}{0.90}
\setlength{\aboverulesep}{1pt}
\setlength{\belowrulesep}{1.5pt}
\begin{threeparttable}
\caption{Storage and relocalization processing cost}
\label{tab:efficiency}
\footnotesize
\setlength{\tabcolsep}{3pt}
\begin{tabular*}{\columnwidth}{@{\extracolsep{\fill}}ccc@{}}
\toprule
Metric & STD-adapt & LP-LIL \\
\midrule
Map cache & 4.19--9.05\,GiB & 167--338\,MiB \\
Serial time/query (ms) & 9835 & 2308 \\
16-core time/query (ms) & 1003 & 183 \\
\bottomrule
\end{tabular*}
\begin{tablenotes}[flushleft]
\footnotesize
\item[] Cache sizes cover the evaluated map and keypoint configurations. Timing excludes startup and includes candidate generation, verification, and refinement. Times were measured over 50-query serial batches and 200-query 16-core batches, and normalized by the batch size minus one. STD-adapt stages were timed sequentially.
\end{tablenotes}
\end{threeparttable}
\end{table}

\subsection{Real-Time Performance}\label{sec:exp-realtime}
With online extrinsic calibration disabled and eight threads for map observations, LP-LIL processed 2838 frames in normal map-constrained localization at a mean cost of 2.494\,ms per frame. Mean costs were 1.502\,ms for map observations, 0.316\,ms for IMU prediction, 0.317\,ms for scan deskewing, 0.173\,ms for voxel downsampling, and 0.041\,ms for IESKF solving. Including pose output, the maximum total was 5.141\,ms, below the 100\,ms interval at 10\,Hz.

Global initialization and relocalization can exceed 100\,ms, but are needed only at startup or to restore map constraints. Once successful, normal map-constrained localization resumes without repeating global search for each scan, supporting real-time localization at 10\,Hz.

\FloatBarrier
\section{Conclusion}\label{sec:conclusion}
In this letter, we presented LP-LIL, a tightly-coupled LiDAR-inertial localization system for navigation across prior-map boundaries. We fuse IMU and scan-to-map measurements using an IESKF and switch to local LIO when prior-map constraints become insufficient. We also proposed a landmark-pair relocalizer that combines pairwise landmark geometry with plane normals to reduce matching ambiguity and enable global localization without an accurate initial pose estimate. Experiments in parking lots, building complexes, and urban roads demonstrate centimeter-level in-map localization accuracy, successful recovery after all 13 departures, and higher global relocalization success rates with lower storage and processing costs than a triangle-descriptor baseline. LP-LIL achieves an average processing time of approximately 2.5 ms per frame, demonstrating its suitability for real-time localization on mobile robots.

\bibliographystyle{IEEEtran}
\bibliography{refs}
\end{document}